\PassOptionsToPackage{numbers}{natbib}

\documentclass{article}

\usepackage[preprint]{style_file}
\pdfoutput=1

\usepackage[utf8]{inputenc} 
\usepackage[T1]{fontenc}    
\usepackage{hyperref}       
\usepackage{url}            
\usepackage{booktabs}       
\usepackage{amsfonts}       
\usepackage{nicefrac}       
\usepackage{microtype}      

\usepackage{amsmath}        
\usepackage{textcomp}      
\usepackage{listings}
\usepackage[table]{xcolor} 
\usepackage{graphicx} 
\usepackage{xspace}
\usepackage{multirow}
\usepackage{booktabs}
\usepackage{wrapfig}
\usepackage{caption}
\usepackage{bbding}
\usepackage{pifont}
\usepackage{tikz}
\title{Generalized Category Discovery in Event-Centric Contexts: Latent Pattern Mining with LLMs} 

\author{
Yi Luo$^{1}$\thanks{Equal contribution.},
Qiwen Wang$^{2}$\footnotemark[1],
Junqi Yang$^{3}$,
Luyao Tang$^{1}$,\\
\textbf{Zhenghao Lin$^{1}$},
\textbf{Zhenzhe Ying$^{3}$},
\textbf{Weiqiang Wang$^{3}$},
\textbf{Chen Lin$^{1,2}$}\thanks{Corresponding author, chenlin@xmu.edu.cn} \\
$^{1}$School of Informatics, Xiamen University \\
$^{2}$Institute of Artificial Intelligence, Xiamen University \\
$^{3}$Ant Group
}

\begin{document}

\newcommand{\method}{\texttt{PaMA}\xspace}
\maketitle

\begin{abstract}
Generalized Category Discovery (GCD) aims to classify both known and novel categories using partially labeled data that contains only known classes. Despite achieving strong performance on existing benchmarks, current textual GCD methods lack sufficient validation in realistic settings. We introduce Event-Centric GCD (EC-GCD), characterized by long, complex narratives and highly imbalanced class distributions, posing two main challenges: (1) divergent clustering versus classification groupings caused by subjective criteria, and (2) Unfair alignment for minority classes. To tackle these, we propose \method, a framework leveraging LLMs to extract and refine event patterns for improved cluster-class alignment. Additionally, a ranking-filtering-mining pipeline ensures balanced representation of prototypes across imbalanced categories. Evaluations on two EC-GCD benchmarks, including a newly constructed Scam Report dataset, demonstrate that \method outperforms prior methods with up to 12.58 \% H-score gains, while maintaining strong generalization on base GCD datasets.
\end{abstract}

\section{Introduction}
Generalized Category Discovery (GCD)~\cite{GCD} seeks to classify samples from both known and novel categories using partially labeled training data. Unlike traditional classification tasks, GCD requires the classifier to go beyond the mere memorization of seen categories and develop generalized reasoning abilities, which are critical under open world uncertainty~\cite{concept_gcd}. Therefore, GCD has attracted considerable research attention from the computer vision and natural language processing communities~\cite{GCD, Happy, MTP, DPN, LOOP}. 

Researchers have developed GCD methods that combine supervised learning on the labeled data with various contrastive learning techniques on the unlabeled data~\cite{MTP, LOOP}. Clustering is also commonly adopted because it assigns pseudo labels to the unlabeled data and provides positive prototypes to the contrastive loss~\cite{TAN, DPN}. Aligning the clusters of unlabeled data with classes of labeled data is considered an essential step to acquire representations that are discriminative for both known and novel categories. The alignment can be achieved by explicitly matching clusters with classes~\cite{DPN}, transferring the prototypes of clusters to classes~\cite{TAN}, or improving the clustering quality using LLM feedback~\cite{LOOP, ALUP, GLEAN}. 


Previous GCD approaches have demonstrated notable progress on textual benchmarks~\cite{bank, stackoverflow, clinc}, which represent relatively simple text classification tasks, such as intent recognition and question classification. However, the effectiveness of previous GCD approaches remains questionable in more complex, real-world GCD scenarios, particularly those involving event-centric narratives~\cite{event-centric}, such as legal cases, clinical reports, and scam reports. We refer to these complex real-world tasks as \textbf{Event-Centric Generalized Category Discovery (EC-GCD)}. Specifically, EC-GCD poses \textit{two severe challenges} to the alignment between clusters and classes. 
\begin{wrapfigure}{r}{0.5\textwidth}
    \centering
    \includegraphics[width=\linewidth]{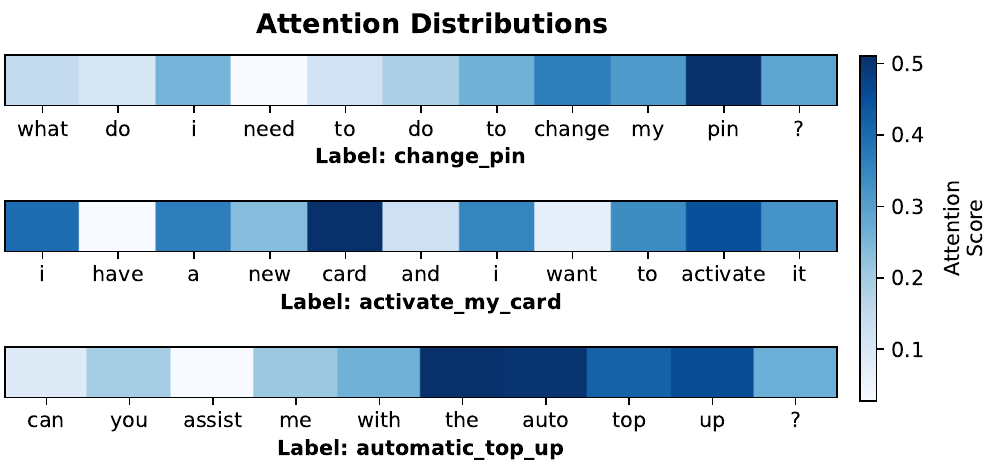}
    \caption{An example of question classification where the model only needs to attend to a few keywords to assign a label accurately.}
    \label{fig:easy_case}
    \vspace{-10pt}
\end{wrapfigure}
Challenge \textbf{\ding{182}}: \textbf{Divergent groupings by clustering and classification}. Classification relies on human-defined labels, whereas clustering is driven solely by similarity of representations. Existing textual GCD benchmarks involve structurally simple short texts, where surface-level cues (e.g., salient keywords) suffice for relatively consistent clustering and classification results (Figure~\ref{fig:easy_case}). In contrast, EC-GCD contains significantly longer texts — our dataset's average length is 14 times that of previous GCD datasets — and features complex logical structures and subjective classification criteria. Consequently, the similarity calculation based on representations may diverge from human annotations. For instance, scam reports of similar event sequences (e.g., victims seek online services and scammers vanish after payment) may be grouped together in the clustering results, whereas human annotators might categorize reports based on service context (e.g., loan applications vs. driving license or exam proxies). Thus, it is challenging to establish an alignment between clusters and classes.

Challenge \textbf{\ding{183}}: \textbf{Unfair alignment for minority classes}. Current benchmarks construct balanced categories, where each category consists of carefully selected samples that reflect predefined class definitions. This ensures clear separations between categories, making it more likely for the clustering algorithm to produce identical groupings as the classes. On the contrary, EC-GCD scenarios exhibit highly imbalanced category distributions. For example, online shopping scams are $100\times$ more frequent than the rarest category in our scam report dataset. The skewed distribution causes dominant-class samples to spread across multiple clusters, dictating the prototype representations of clusters that should correspond to minority classes. Meanwhile, minority-class instances, due to their lower representation in feature space, are at risk of being incorrectly merged into dominant clusters. 

We propose \method (\textbf{Pa}ttern \textbf{M}ining and \textbf{A}lignment), a framework that is tailored for EC-GCD tasks. To address Challenge \textbf{\ding{182}}, \method leverages LLMs to improve semantic understanding of complex event-centric narratives. The LLM extracts event patterns for each cluster, which are then refined using labeled data to align with human classification standards. The event patterns are utilized to adjust cluster assignments and cluster prototypes. Thus, LLM provides additional supervision to enhance the accuracy of cluster-class alignment. To address Challenge \textbf{\ding{183}}, \method ranks clusters by size and compactness and then summarizes event patterns in this ranked order. This strategy ensures that dominant-class patterns are identified first, without overshadowing those from minority classes, thereby promoting diversity in cluster prototypes. As a result, rare categories and new classes are effectively separated into distinct clusters.

We evaluate \method on two real-world EC-GCD datasets, including a newly constructed benchmark, \textit{Scam Report}. \method consistently outperforms all baselines on EC-GCD, with up to a \textbf{12.58\%} gain in H-score over the strongest competitor. To assess generalization, we further evaluate on three standard GCD benchmarks, where \method achieves competitive performance and maintains a strong balance between known and novel class recognition.


In summary, our work makes the following \textbf{contributions}:
\setlength{\itemindent}{20pt}
\setlength{\leftmargini}{20pt}
\begin{enumerate}
\item We propose a set of LLM-guided adjustment strategies to enhance alignment between labeled and unlabeled data for event-centric GCD.

\item We propose a ranking-filtering-mining pipeline that improves alignment for minority classes in imbalanced GCD settings.

\item We build a new real-world EC-GCD benchmark, \textit{Scam Report}, and demonstrate state-of-the-art performance on two EC-GCD datasets.
\end{enumerate}

\section{Related Work}
\paragraph{Generalized Category Discovery.}
Generalized Category Discovery (GCD)~\cite{GCD, NCL, DeCrisisMB, MTP, mou} has emerged as a key problem under the open-world setting, where models are trained on a small amount of labeled data with known categories to recognize both known and novel categories within unlabeled data. Early works in the textual domain~\cite{MTP, DPN, KTN} typically follow a three-stage process: pretraining with mixed supervised and unsupervised data, self-supervised learning on unlabeled data, and clustering methods such as K-Means. To reduce bias toward known classes, Wen et al.~\cite{SimGCD}, Bai et al.~\cite{BaCon}, and An et al.\cite{DPN} introduced soft pseudo-labels and parametric prototypes, while le Shi et al.~\cite{KTN} and An et al. \cite{TAN} applied knowledge transfer to connect known and novel categories. However, the ambiguous boundaries and noisy pseudo-labels in novel classes remain challenges, prompting recent research to incorporate large language models (LLMs) to improve category discovery.

\paragraph{LLM-based Active Learning in GCD.}
Active Learning (AL) selectively identifies the most informative samples for annotation, improving learning efficiency~\cite{karamcheti, MHPL}. AL methods can be categorized into uncertainty-based methods~\cite{pmlr, margatin},  diversity-based methods~\cite{Citovsky, ACTUNE}, and hybrid methods~\cite{DeepAL+}. In GCD, An et al.~\cite{LOOP}, Liang et al.~\cite{ALUP}, and Zou et al.\cite{GLEAN} leverage LLMs to evaluate relationships between samples or refine pseudo-labels, providing additional supervision to enhance representation learning. Furthermore, Zou et al.\cite{GLEAN} use LLMs to generate cluster-level category descriptions and align samples to these descriptions. However, they do not explicitly align feature spaces between labeled and unlabeled data. Despite the achievement in conventional GCD, many real-world GCD tasks are event-centric, where category distinctions emerge from latent patterns in event-centric texts. This setting introduces two main challenges: (1) Divergent groupings by clustering and classification; (2) Unfair alignment for minority classes. Existing methods \cite{probabilistic, DPN, CsePL, KTN, TAN, LOOP, ALUP, GLEAN} are primarily optimized on simpler textual domains and have yet to fully address these challenges, which motivates the focus of our work.

\section{Method}
\paragraph{Problem Formulation}

Event-Centric Generalized Category Discovery (EC-GCD) follows the same core formulation as Generalized Category Discovery (GCD)~\cite{GCD}. Given a labeled dataset $D_l = \{(x_i, y_i) \mid y_i \in Y_k\}$ with known categories $Y_k$, models are expected to generalize to unlabeled data $D_u = \{x_i \mid y_i \in \{Y_k, Y_n\} \}$ containing both known and novel categories $Y_n$. The objective is to jointly recognize all categories in the combined training set $D_{\text{all}} = D_l \cup D_u$, with evaluation on a test set $D_t = \{(x_i, y_i) \mid y_i \in \{Y_k, Y_n\}\}$. The key difference in EC-GCD lies in its domain: event-centric texts where successful classification depends on identifying latent patterns rather than surface-level cues. Following prior work~~\cite{GCD, LOOP}, we assume the total number of categories $K$ is known.

\paragraph{Overview}

\begin{figure}[t]
  \centering
  \includegraphics[width=\textwidth]{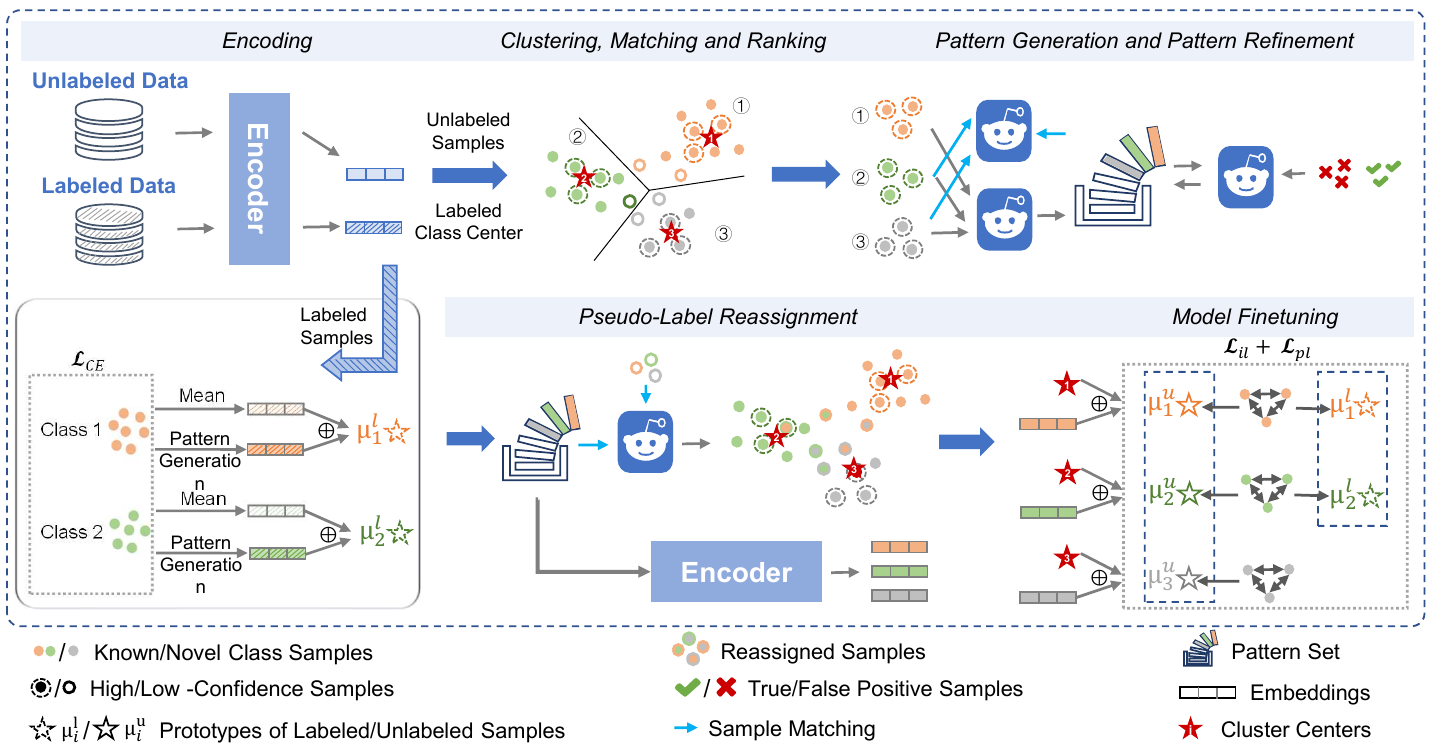}
  \caption{The framework of \method. The pretraining stage is omitted for clarity.}
  \label{fig:framework}
\end{figure}

\method comprises two stages: encoder pretraining, following prior work~\cite{MTP, DPN, GLEAN}, and fine-tuning (Figure~\ref{fig:framework}). Each fine-tuning epoch begins by encoding the input data. Unlabeled embeddings are clustered in an unsupervised manner, and cluster centers are matched to labeled class centroids to identify known and novel classes. Clusters are then ranked by size and compactness (Sec. ~\ref{sec:cluster_rank}). After ranking, an LLM generates representative patterns for each cluster in order of priority, which are then refined with labeled data (Sec.~\ref{sec:pattern_generate_refine}). Low-confidence samples are reassigned pseudo-labels based on their alignment with the patterns (Sec.~\ref{sec:psudo-label_reassignment}). Finally, the encoder is trained to minimize intra-class variance and enhance instance-prototype alignment (Sec.~\ref{sec:loss}).


\subsection{Clustering, Matching and Ranking}
\label{sec:cluster_rank}
Unsupervised clustering often misaligns with class boundaries, particularly under class imbalance, where dominant classes may span multiple clusters. This leads to redundant patterns across different clusters and hinders prototype learning. To address this, we introduce a cluster ranking mechanism based on normalized compactness and size, enabling early capture of dominant class semantics while reducing their influence on subsequent pattern generation. Specifically, we apply K-means to the unlabeled data embeddings, assuming a known number of categories $K$, as in previous work~\cite{GCD, LOOP}. In open-world settings where $K$ is unknown, we can adapt the estimation strategy from Vaze et al.~\cite{GCD}. Following the Alignment and Decoupling framework in DPN~\cite{DPN}, which assumes proximate centers in feature space correspond to identical categories, we can find the optimal match function \(P(\cdot)\) through the Hungarian algorithm~\cite{Hungarian}. \(\mu^l_i\) and \(\mu^u_{P(i)}\) represent matched known-class centers, while unmatched clusters are treated as novel categories.

For each cluster \(C_k\), we compute two normalized scores: \textbf{Compactness}, which favors low intra-cluster variance, and \textbf{Size}, which favors larger clusters likely representing high-frequency classes:

\begin{minipage}{0.5\linewidth}
\begin{equation}
    \text{Comp.}_k = 
\begin{cases}
\frac{\max_d - d_k}{\max_d - \min_d}, & \text{if } \max_d \ne \min_d, \\
1, & \text{otherwise},
\end{cases}
\label{eq:comp}
\end{equation}
\end{minipage}
\hfill
\begin{minipage}{0.5\linewidth}
\vspace{-10pt}
\begin{equation}
    \text{Size}_k = 
    \begin{cases}\frac{s_k - \min_s}{\max_s - \min_s}, & \text{if } \max_s \ne \min_s \\ 1, & \text{otherwise}.
    \end{cases}
\label{eq:size}
\end{equation}
\end{minipage}

Here, \(d_k\) denotes average intra-cluster distance, and \(s_k\) is the cluster size; \(\max_d, \min_d\) and \(\max_s, \min_s\) are the respective extremes across clusters.

Each cluster is assigned an overall ranking score via a weighted sum:
\begin{equation}
    \text{Score}(C_k) = \sigma \cdot \text{Comp.}_k + (1 - \sigma) \cdot \text{Size}_k,
\end{equation}
where \(\sigma \in [0, 1]\) controls the trade-off between reliability and dominance.

This ranking mechanism prioritizes well-formed, dominant clusters in the pattern generation and refinement stage (Sec.~\ref{sec:pattern_generate_refine}), improving both pattern reliability and diversity.

\subsection{Pattern Generation and Refinement}
\label{sec:pattern_generate_refine}
Surface-level cues are often inadequate for characterizing categories in EC-GCD tasks. To address this, we leverage an LLM to uncover latent patterns in the data for accurate classification.

\subsubsection{High-Confidence Sample Selection}
\label{sec:high-confidence}
To optimize both computational efficiency and pattern quality, we employ a \textit{high-confidence sample selection} approach to mitigate intra-cluster feature conflation, where excessive diversity within clusters leads to noisy patterns by merging discriminative features across categories. 

\textbf{Selection Criteria.} Each sample \(x_i\) is evaluated through two complementary metrics:
1. Predictive Uncertainty~\cite{ALUP, LOOP}: Quantified by the entropy \( \mathcal{H}_i \) of its cluster assignment distribution \( q_i \).  
2. Centrality: Measured by the Euclidean distance \( \|f_\theta(x_i) - \mu_j\|_2 \) to its assigned cluster center \( \mu_j \).
We formulate the assignment probability \( q_{ij} \) using a Student's \textit{t}-distribution~\cite{student-t} as in Eq.~\ref{eq:q}. The predictive uncertainty is then derived as in Eq.~\ref{eq:h}:

\begin{minipage}{0.5\linewidth}
\begin{equation}
q_{ij} = \frac{\left(1 + \|f_\theta(x_i) - \mu_j\|^2 / \alpha \right)^{-\frac{\alpha+1}{2}}}
{\sum_{j'} \left(1 + \|f_\theta(x_i) - \mu_{j'}\|^2 / \alpha \right)^{-\frac{\alpha+1}{2}}},
\label{eq:q}
\end{equation}
\end{minipage}
\hfill
\begin{minipage}{0.5\linewidth}
\vspace{17pt}
\begin{equation}
    \mathcal{H}_i = -\sum_j q_{ij} \log q_{ij}.
\label{eq:h}
\end{equation}
\end{minipage}

where \( f_\theta(\cdot) \) denotes the feature encoder, $\mu_j$ is the center of cluster $j$, and \(\alpha\) is the degrees of freedom. 

Finally, samples are selected by first intersecting the top-k highest-centrality and highest-certainty candidates. Remaining slots are filled sequentially from the distance-ranked candidates, followed by entropy-ranked candidates until k samples are obtained.

\subsubsection{Pattern Generation} 
\label{sec:pattern_generate}
While high-confidence sample selection helps reduce noise, outliers may still occur. To maintain pattern quality, we implement a filtering-then-generating process (Figure~\ref{fig:pattern}(a)): 

\begin{figure}[h]
  \centering
  \includegraphics[width=\textwidth]{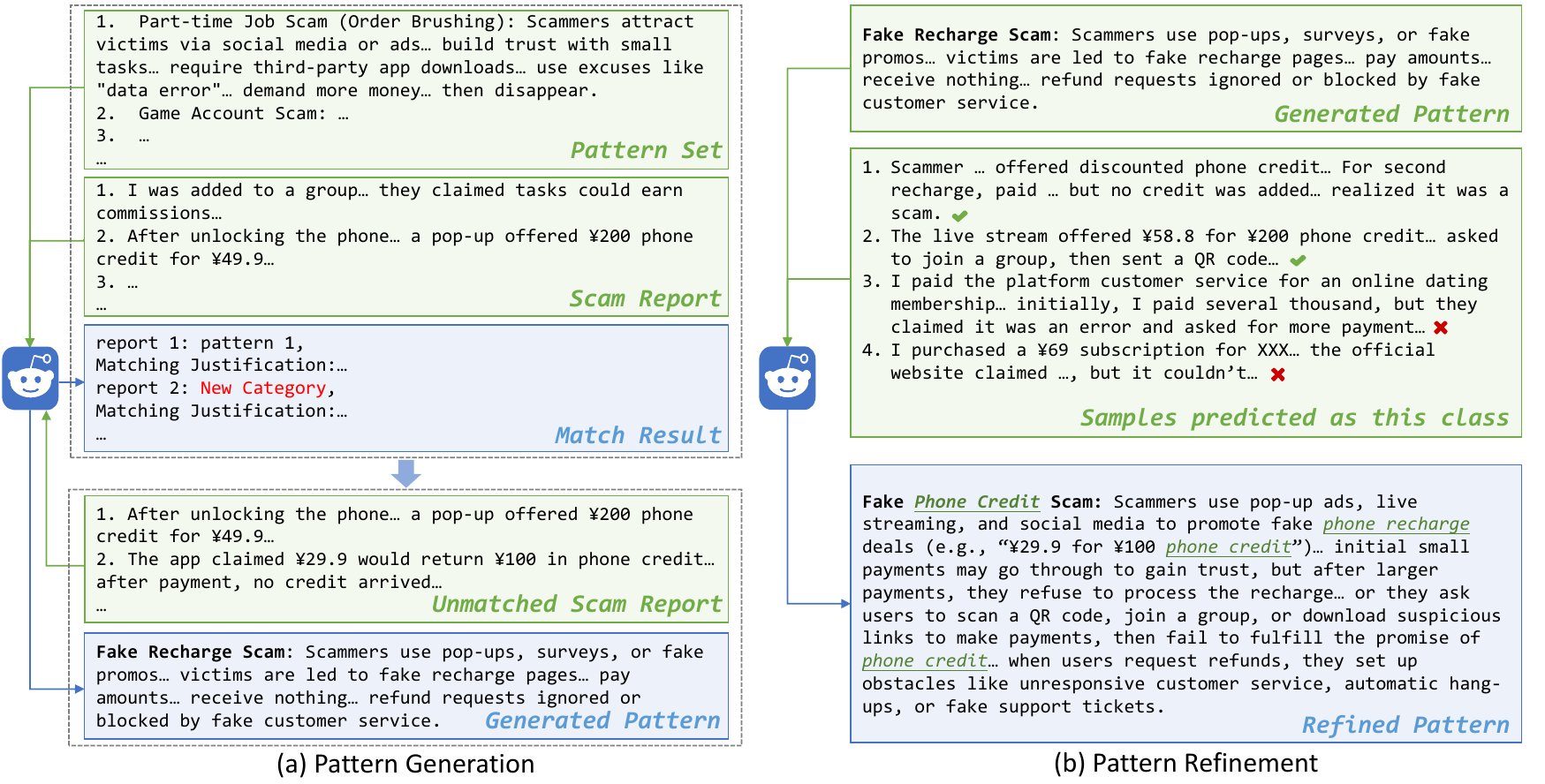}
  \caption{An example of Pattern Generation and Refinement. (a) Top: Sample Matching. Bottom: Consensus-Driven Extraction. (b) Pattern Refinement example.}
  \label{fig:pattern}
\end{figure}
\vspace{-10pt}

1. \textbf{Sample Matching}: As shown in the upper part of Figure~\ref{fig:pattern}(a), each sample is first compared against existing patterns in the current pattern set (initially empty for each new cluster). A detailed example is shown in Appendix~\ref{sec:example_matching}. Samples matching existing patterns are filtered out. 

2. \textbf{Consensus-Driven Extraction}: For unmatched samples (i.e., labeled as \texttt{New Category}), the LLM is prompted to identify a dominant underlying category and generate a representative pattern accordingly. Crucially, the prompt constrains the LLM to exclude mismatched samples, ensuring that the generated pattern remains category-specific. A detailed example is shown in Appendix~\ref{sec:example_generation}.

This approach not only reduces the impact of anomalous data but also ensures that generated patterns accurately reflect the core features within the cluster.

\subsubsection{Pattern Refinement}
Classification boundaries in EC-GCD are subjective and shaped by annotator bias. Relying solely on LLM knowledge can misalign patterns with human-labeled standards. To mitigate this, we introduce a refinement process that leverages both positive and negative learning signals. Specifically, each pattern is refined based on: (1) the original pattern, and (2) labeled samples predicted as its class, including both true positives and false positives. The LLM revises the pattern to preserve correct associations with true positives while excluding false positives. As illustrated in Figure~\ref{fig:pattern}(b), the initial pattern "fake recharge scam" is refined after observing that samples referring to fake membership recharges were incorrectly matched, obtaining the revised pattern focusing on phone credit scams. This refinement step explicitly encodes human-defined classification criteria into the patterns, improving alignment with known-class decision boundaries.

\subsection{Pseudo-Label Reassignment}
\label{sec:psudo-label_reassignment}
Unsupervised clustering often yields noisy pseudo-labels that hinder model training. To address this, we design a \textbf{Pseudo-Label Reassignment} process. We first identify low-confidence samples based on two criteria: (1) the uncertainty score used during high-confidence sample selection (Sec.~\ref{sec:high-confidence}), and (2) a novel stability-based metric that captures clustering inconsistency-samples that frequently switch cluster assignments across multiple runs are deemed unstable. The final low-confidence set is defined as the intersection of these unstable samples and the top-$k$ highest-entropy instances.

Given the low-confidence set, our reassignment procedure comprises two main steps: (1) low-confidence samples and non-majority instances (identified via the \textit{consensus-based extraction} step in Sec.~\ref{sec:pattern_generate}) are re-evaluated and assigned new pseudo-labels based on their best-matching refined patterns; (2) Samples matched to existing patterns during the pattern generation phase are also relabeled accordingly. This process mitigates the impact of noisy pseudo-labels and improves classification performance.

\subsection{Optimization Objectives}
\label{sec:loss}
\paragraph{Instance-Level Contrastive Learning}
We apply instance-level contrastive learning to align samples within the same cluster and separate those from different clusters. Specifically, we use the InfoNCE loss~\cite{InfoNCE} over samples processed by the LLM, including both confirmed high-confidence samples and those with reassigned pseudo-labels. For each anchor sample $x_i$, a positive $x_j$ is sampled from the same cluster, while $N$ negatives are drawn from other clusters. The loss is defined as:

\vspace{-5pt}
\begin{equation}
\mathcal{L}_{\text{il}} = -\frac{1}{n} \sum_{i=1}^{n}\log \frac{\exp(\mathrm{sim}(f_{\theta}(x_i), f_{\theta}(x_j)/\tau)}{\sum_{k=1}^{N+1} \exp(\mathrm{sim}(f_{\theta}(x_i), f_{\theta}(x_k)/\tau)}.
\label{eq:infoNCE}
\end{equation}
\vspace{-5pt}

Here, $\{\mathbf{x}_k\}_{k=1}^{N+1}$ consists of one positive and N negatives sampled from distinct clusters, $\mathrm{sim}(\cdot, \cdot)$ denotes cosine similarity, and $\tau$ is a temperature hyperparameter.

\paragraph{Hybrid Prototype Learning}
While instance-level objectives capture local alignment, they often overlook the higher-level semantic structure between instances and their categories. To address this, we incorporate prototypical learning (PL)~\cite{prototype-learning}, which promotes alignment between samples and their corresponding cluster prototypes while repelling them from unrelated prototypes. We employ a prototype-level contrastive loss similar to instance-level contrastive learning. For samples whose pseudo-labels have changed during reassignment, a weighting factor $\rho$ is applied:

\vspace{-5pt}
\begin{equation}
\label{eq:pl}
\mathcal{L}_{\text{pl}} = -\frac{1}{n} \sum_{i=1}^{n} w_i \cdot \log \frac{\exp\left(\mathrm{sim}(f_{\theta}(x_i), P_j)/\tau\right)}{\sum_{k=1}^{N+1} \exp\left(\mathrm{sim}(f_{\theta}(x_i), P_k)/\tau\right)},
\end{equation}

\begin{equation}
w_i = 
\begin{cases}
\rho, & \text{if the pseudo-label of } x_i \text{ has changed}, \\
1, & \text{otherwise},
\end{cases}
\end{equation}
\vspace{-5pt}

where $x_i$ belongs to cluster $C_j$, and $P_j$ is its corresponding prototype.

Following An et al.~\cite{DPN}, we apply distinct PL objectives for known and novel categories. For novel categories, we encourage alignment between unlabeled samples and their cluster prototypes (Eq.~\ref{eq:novel}). For known categories, we promote consistency between unlabeled samples and both unlabeled and labeled prototypes, facilitating knowledge transfer from labeled data (Eq.~\ref{eq:known}). 

\vspace{-5pt}
\begin{minipage}{0.5\linewidth}
\begin{equation}
    \mathcal{L}_{\text{novel\_pl}} = \mathcal{L}_{\text{pl}}(\mathcal{D}^{un}, \mathcal{P}^{un}),
\label{eq:novel}
\end{equation}
\end{minipage}
\hfill
\begin{minipage}{0.5\linewidth}
\vspace{0pt}
\begin{equation}
    \mathcal{L}_{\text{known\_pl}} = \mathcal{L}_{\text{pl}}({D}^{uk}, {P}^{uk}) + \mathcal{L}_{\text{pl}}({D}^{uk}, {P}^{l}).
\label{eq:known}
\end{equation}
\end{minipage}
\vspace{-5pt}

In Eq.~\ref{eq:novel}, \({D}^{un}\) and \({P}^{un}\) denote the data and prototypes for novel categories, respectively. In Eq.~\ref{eq:known}, \({D}^{uk}\) and \({P}^{uk}\)denote the data and prototypes for known categories in the unlabeled set, and \({P}^l\) represent the labeled data's prototypes. 

\paragraph{Prototype Calculation and Update.}
We define prototypes by integrating both statistical and semantic cues. For each class  $C_i$, the prototype $P_i$ is computed as a weighted combination of its class center $\mu_i$ and the embedding of its representative pattern $p_i$, with $\beta$ controling the balance between statistical and semantic information (Eq.~\ref{eq:p_i}). We further stabilize training by updating prototypes via exponential moving average (Eq.~\ref{eq:p_t}).

\begin{minipage}{0.5\linewidth}
\begin{equation}
P_i = \beta\mu_i + (1-\beta)f_\theta(p_i),
\label{eq:p_i}
\end{equation}
\end{minipage}
\hfill
\begin{minipage}{0.5\linewidth}
\vspace{0pt}
\begin{equation}
P^{(t+1)} \leftarrow \omega P^{(t)} + (1 - \omega) {P}^{(t+1)},
\label{eq:p_t}
\end{equation}
\end{minipage}

where \(\omega\) is a momentum coefficient, and \({P}^{(t)}\) is the prototype computed at epoch $t$. This smoothing improves prototype stability over time.

\paragraph{Total Loss.} To prevent catastrophic forgetting, we include a cross-entropy loss on labeled data ($\mathcal{L}_{\text{ce}}(\mathcal{D}^l)$). The overall training objective is:

\vspace{-5pt}
\begin{equation}
    \mathcal{L}_{\text{\method}} = \mathcal{L}_{\text{il}} + \mathcal{L}_{\text{novel\_pl}} + \mathcal{L}_{\text{known\_pl}} + \mathcal{L}_{\text{ce}}.
\end{equation}
\section{Experiment}
\label{sec:experiment}
\subsection{Experiment Setup}
\label{sec:setup}
\paragraph{Datasets}
We evaluate our method on two Chinese EC-GCD datasets and three English base GCD benchmarks. The first, \textbf{Scam Report}, is a crowdsourced annotated dataset constructed by our team. The second, \textbf{Telecom Fraud Case}~\cite{ccl}, consists of police transcripts from real fraud cases\footnote{Released by the CCL2023 Evaluation organized by Harbin Institute of Technology.}. The three conventional GCD datasets include: \textbf{BANKING} (banking-domain intent detection,~\cite{bank}), \textbf{StackOverflow} (question classification,~\cite{stackoverflow}), and \textbf{CLINC} (multi-domain intent detection~\cite{clinc}). Dataset statistics are provided in Table~\ref{tab:dataset_stats} of the Appendix~\ref{sec:implement}.

\paragraph{Baselines} We compare our model with the following methods: (1) \textbf{MTP}~\cite{MTP} conducts multi-task pretraining with masked language modeling (MLM) and classification (the subsequent work continues this pretraining approach), followed by reducing intra-cluster pairwise distances; (2) \textbf{DPN}~\cite{DPN} aggregates samples toward all prototypes using distance-weighted averaging; (3) \textbf{TAN}~\cite{TAN} aligns labeled and unlabeled prototypes while enforcing instance–prototype and intra-class consistency; (4) \textbf{ALUP}~\cite{ALUP} moves ambiguous samples toward cluster-representative instances; (5) \textbf{LOOP}~\cite{LOOP} pulls ambiguous samples toward correct neighbors; (6) \textbf{GLEAN}~\cite{GLEAN} improves LOOP by generating class descriptions to align ambiguous instances with cluster-level semantic information.

\paragraph{Evaluation Metrics} Following prior work~\cite{LOOP}, we evaluate model performance using clustering accuracy, computed by aligning predicted cluster labels with ground-truth labels via the Hungarian algorithm~\cite{Hungarian}. We report: (1) \textbf{$\text{ACC}_\text{K}$}, the accuracy on known categories; (2) \textbf{$\text{ACC}_\text{N}$}, the accuracy on novel categories; and (3) \textbf{H-score}~\cite{h-score}, the harmonic mean of accuracy on known and novel categories, the detailed formulation is provided in Appendix~\ref{sec:implement}. 

\paragraph{Implementation Details}
We use BERT~\cite{BERT} as the backbone encoder, with the final-layer \texttt{[CLS]} token serving as the instance representation. For the LLM component, we adapt Qwen2.5-72B-Instruct-AWQ\footnote{\url{https://huggingface.co/Qwen/Qwen2.5-72B-Instruct-AWQ}}~\cite{qwen2.5}. Baselines are implemented based on their original designs and reported hyperparameters, with the LLM replaced for consistency. Full hyperparameter settings are listed in Table~\ref{tab:hyperparameter} (Appendix~\ref{sec:implement}). In our main experiments, we use the optimal sample weight of $\rho = 25$ for the scam report, while the ablations use the initial value of $\rho = 1$. We report results averaged over three runs. To reduce overhead, we update the clustering and invoke the LLM every five epochs.

\subsection{Main Results}
\label{sec:result}
\setlength{\tabcolsep}{3pt}
\definecolor{darkgreen}{rgb}{0.0, 0.5, 0.0}
\definecolor{darkred}{rgb}{0.55, 0.0, 0.0} 

\begin{table}[ht]
\caption{Results (\%) of \method compared to baseline methods across different datasets. Some results are cited from An et al.~\cite{LOOP, TAN}. The results on Scam Report are obtained using the optimal sample weight $\rho=25$, while the initial $\rho$ value of 1 is used for other datasets.}
\label{tab:main_result}
\resizebox{\textwidth}{!}{
\begin{tabular}{ccccccccc>{\columncolor{gray!10}}l}
\toprule
Task &Dataset                                & Method  & MTP   & DPN   & TAN   & ALUP  & LOOP  & GLEAN & \method \textbf{(Ours)}  \\\cmidrule(lr){1-1} \cmidrule(lr){2-2} \cmidrule(lr){3-10}
\multirow{6}{*}{EC-GCD} & \multirow{3}{*}{Scam Report}        & $\text{ACC}_\text{K}$  & 28.28 & 30.56 & 46.52 & 40.48 & 40.02 & 37.17 & \textbf{60.04} \textcolor{darkgreen}{(+13.52)} \\
                        &                                     & $\text{ACC}_\text{N}$  & 36.18 & 17.59 & 32.16 & 32.16 & 22.61 & 32.66 & \textbf{44.10} \textcolor{darkgreen}{(+7.92)}  \\
                        &                                     & H-score & 31.70 & 22.32 & 38.03 & 35.84 & 28.89 & 34.77 & \textbf{50.88} \textcolor{darkgreen}{(+12.58)} \\\cmidrule(lr){2-2} \cmidrule(lr){3-10}
                        & \multirow{3}{*}{Telecom Fraud Case} & $\text{ACC}_\text{K}$  & 55.43 & 79.59 & 71.19 & 74.29 & 57.62 & 52.71 & \textbf{83.18} \textcolor{darkgreen}{(+3.59)}  \\
                        &                                     & $\text{ACC}_\text{N}$  & 43.40 & 45.28 & 57.55 & 42.45 & 28.30 & 65.57 & \textbf{66.67} \textcolor{darkgreen}{(+1.10)}  \\
                        &                                     & H-score & 48.68 & 57.72 & 63.65 & 54.02 & 37.96 & 58.44 & \textbf{74.05} \textcolor{darkgreen}{(+10.40)}  \\\midrule
\multirow{9}{*}{Base GCD}    & \multirow{3}{*}{BANKING}            & $\text{ACC}_\text{K}$  & 80.08 & 80.93 & 81.97 & 74.09 & \textbf{84.78} & 75.15 & 72.80 \textcolor{darkred}{(-11.98)}  \\
                        &                                     & $\text{ACC}_\text{N}$  & 50.04 & 48.60 & 56.23 & 46.05 & 60.13 & 73.06 & \textbf{78.16} \textcolor{darkgreen}{(+5.10)}  \\
                        &                                     & H-score & 61.59 & 60.73 & 66.70 & 56.80 & 70.35 & 74.09 & \textbf{75.38} \textcolor{darkgreen}{(+1.29)} \\\cmidrule(lr){2-2} \cmidrule(lr){3-10}
                        & \multirow{3}{*}{StackOverflow}      & $\text{ACC}_\text{K}$  & 84.75 & 85.29 & \textbf{86.36} & 52.00 & 84.13 & 84.93 & 78.00 \textcolor{darkred}{(-8.36)} \\
                        &                                     & $\text{ACC}_\text{N}$  & 70.93 & 81.07 & 86.93 & 45.38 & 86.40 & \textbf{91.20} & 86.00 \textcolor{darkred}{(-5.20)}  \\
                        &                                     & H-score & 77.23 & 83.13 & 86.64 & 40.27 & 85.25 & \textbf{87.96} & 81.80 \textcolor{darkred}{(-6.16)}  \\\cmidrule(lr){2-2} \cmidrule(lr){3-10}
                        & \multirow{3}{*}{CLINC}              & $\text{ACC}_\text{K}$  & 91.69 & 92.97 & \textbf{93.39} & 90.12 & 92.08 & 83.95 & 89.70 \textcolor{darkred}{(-3.69)}  \\
                        &                                     & $\text{ACC}_\text{N}$  & 71.46 & 77.54 & 81.46 & \textbf{84.91} & 75.61 & 74.23 & \textbf{84.91} \textcolor{darkgreen}{(+0.00)}  \\
                        &                                     & H-score & 80.32 & 84.56 & 87.02 & \textbf{87.43} & 83.04 & 78.79 & 87.24 \textcolor{darkred}{(-0.19)}  \\\bottomrule
\end{tabular}
}
\end{table}


Table~\ref{tab:main_result} reports the performance of \method compared to several strong baselines across five datasets under two GCD tasks: EC-GCD and Base GCD. The results demonstrate three key advantages of our approach:
\textbf{(1) Superior performance on challenging EC-GCD tasks.} On Scam Report, \method achieves the highest H-score, surpassing the best baseline by \textbf{12.58\%}. Its $\text{ACC}_\text{K}$ reaches \textbf{1.62×} the baseline average. On Telecom Fraud Case, our method again outperforms all baselines, with a \textbf{10.40\%} improvement in H-score over the best competitor.
\textbf{(2) Comparable Performance on Base GCD Benchmarks.} Our method achieves the highest $\text{ACC}_\text{N}$ and H-score on BANKING, ties for best $\text{ACC}_\text{N}$ on CLINC with competitive H-scores. Although our $\text{ACC}_\text{K}$ on BANKING is 11.98\% lower than the best competitor LOOP, our $\text{ACC}_\text{N}$ surpasses it by \textbf{18.03\%}. This balanced performance in both metrics results in the highest H-score. On StackOverflow, a modest H-score drop suggests slight limitations in simpler settings. Nonetheless, the overall performance confirms that our design generalizes well beyond EC-GCD.
\textbf{(3) Strong novel-class recognition with balanced known/novel trade-off.} Our method achieves the highest $\text{ACC}_\text{N}$ on \textbf{4 of 5} datasets, including 44.10\% on Scam Report (+7.92\% over MTP), demonstrating strong novel-class detection. It maintains balanced performance across known and novel classes, achieving top H-scores on multiple datasets. This balance reflects our method’s ability to learn discriminative criteria from labeled data and mitigate bias from class imbalance.
In sum, \method effectively tackles the core challenges of EC-GCD: subjectivity, imbalance, and input complexity, while preserving strong generalization on base GCD tasks.

\subsection{Ablation Study}

\subsubsection{Effect of Pattern Refinement}
\begin{figure}[htbp]
    \centering
    \vspace{-20pt}
\begin{minipage}{0.42\linewidth}
\vspace{20pt}
\centering
\small
\captionof{table}{The results (\%) of \method without including different components.}
\label{tab:module_ablation}
\begin{tabular}{lccc}
\toprule
  & $\text{ACC}_\text{K}$ & $\text{ACC}_\text{N}$ & H-score  \\ \midrule
\rowcolor{gray!10}Full Method                                           &  \textbf{63.62}  &  \textbf{33.16}  &   \textbf{43.59}      \\
w/o Refinement                                         &   59.05   &  30.72 &   40.41      \\
w/o Ranking \& Filtering                                      &     62.62   &  28.80 &   39.45       \\\midrule
w/o   $\mathcal{L}_{\text{il}}$                                  &     47.91	&30.54	&37.30        \\
w/o   $\mathcal{L}_{\text{pl}}(\mathcal{D}^{uk}, \mathcal{P}^l)$ &       61.63	&30.89	&41.15     \\
w/o   $\mathcal{L}_{\text{pl}}$                                  &       61.23&	27.05&	37.52        \\
w/o   $\mathcal{L}_{\text{ce}}$                                  &     58.85	&30.54	&40.21   \\ \bottomrule
\end{tabular}
\end{minipage}
\hfill
\begin{minipage}{0.49\linewidth}
  \centering
  \includegraphics[width=\textwidth]{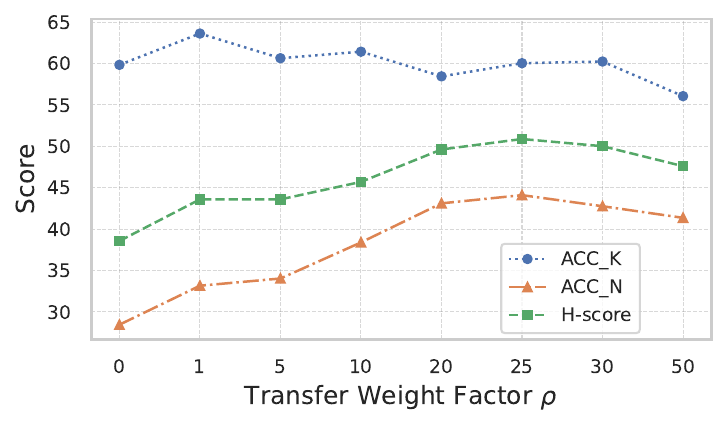}
  \caption{Effect of the transfer weight factor $\rho$.}
    \vspace{-20pt}
  \label{fig:transfer_weight}
\end{minipage}
\vspace{-10pt}
\end{figure}

We evaluate the impact of the pattern refinement module on our Scam Report dataset. As shown in Table~\ref{tab:module_ablation}, removing refinement leads to a drop of 4.57\% in $\text{ACC}_\text{K}$ and 2.44\% in $\text{ACC}_\text{N}$. This demonstrates that refinement effectively aligns generated patterns with human-labeled semantics, improving known class recognition and overall categorization.

\subsubsection{Effect of Cluster Ranking and Sample Filtering}
To assess the role of cluster ranking and sample filtering, we ablate these components from our method. As shown in Table~\ref{tab:module_ablation}, this variant suffers a decline in $\text{ACC}_\text{N}$ (–4.36\%) and H-score (–4.14\%), highlighting the importance of cluster prioritization and noise reduction. These mechanisms are especially beneficial for novel class discovery under class imbalance.

\subsubsection{Effect of Various Optimization Objectives}
To assess the contribution of each optimization objective, we measure performance degradation after removing: (1) the instance-level contrastive loss $\mathcal{L}_{\text{il}}$, (2) the knowledge transfer term in $\mathcal{L}_{\text{known\_pl}}$, i.e., $\mathcal{L}_{\text{pl}}({D}^{uk}, {P}^l)$, (3) the overall prototype learning loss $\mathcal{L}_{\text{pl}}$, and (4) the cross-entropy loss $\mathcal{L}_{\text{ce}}$ for labeled data. As shown in Table~\ref{tab:module_ablation}, excluding $\mathcal{L}_{\text{il}}$ results in the largest decline in $\text{ACC}_\text{K}$ (–15.71\%), highlighting the importance of fine-grained instance discrimination. Removing $\mathcal{L}_{\text{ce}}$ also leads to notable degradation, suggesting its effectiveness in mitigating catastrophic forgetting through supervised learning. Ablating the knowledge transfer term $\mathcal{L}_{\text{pl}}({D}^{uk}, {P}^l)$ reduces both $\text{ACC}_\text{K}$ and $\text{ACC}_\text{N}$. More significantly, fully ablating $\mathcal{L}_{\text{pl}}$ results in the substantial performance drop, particularly in novel class generalization, yielding the lowest $\text{ACC}_\text{N}$ (27.05\%). These results confirm that prototype-level alignment is critical for transferring knowledge from known to unknown classes.

\subsubsection{The Influence of different parameters}
\paragraph{The Training Weight of Reasigned Pseudo-Label samples.}

We conducted an ablation study to assess the impact of the training weight $\rho$ for reassigned pseudo-labeled samples, as shown in Figure~\ref{fig:transfer_weight}. The baseline $\rho=0$ excludes these samples from prototype learning. Varying $\rho$ reveals three key trends: 1) As $\rho$ increases from $0$ to $25$, both $\text{ACC}_\text{N}$ and H-score improve, indicating that incorporating higher-confidence pseudo-labels enhances novel-class prototypes. 2) Performance peaks at $\rho = 25$, with the highest H-score (50.88\%) and $\text{ACC}_\text{N}$ (44.10\%). 3) When $\rho > 25$ (e.g., 30 or 50), performance begins to decline, particularly on known categories. This is likely due to residual noise in pseudo-labels — excessive weighting amplifies incorrect supervision signals.

\paragraph{Known Class Ratio.}

\vspace{-5pt}
\begin{figure}[h]
  \centering
  \includegraphics[width=1.0\textwidth]{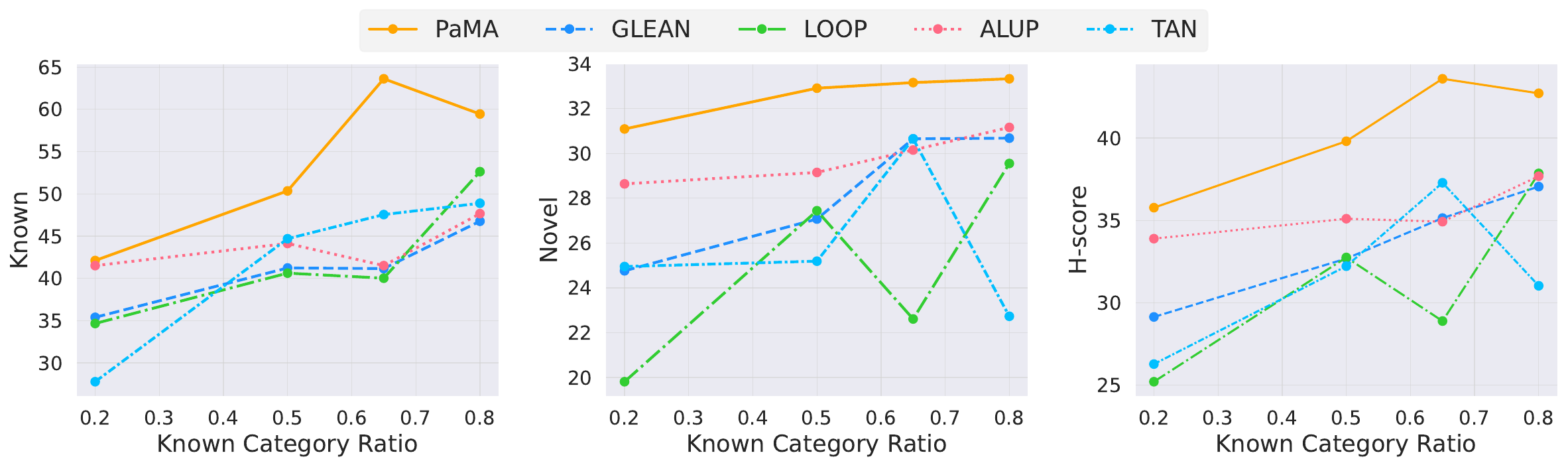}
  \caption{Results (\%) of \method compared to baseline methods across different known class ratios.}
  \label{fig:kcr}
\end{figure}

In real-world scenarios, the proportion of known categories can vary widely. Such dynamic variations impose higher demands on the robustness and adaptability of models. To comprehensively evaluate \method's performance under different proportions of known categories, we design the experiments on the Scam Report dataset, setting four distinct proportions of known categories: $\{0.2, 0.5, 0.65,  0.8\}$. As shown in Figure \ref{fig:kcr}, regardless of the proportion of known categories, \method significantly outperforms other comparative methods across all evaluation metrics. Notably, even in the extreme case where the known class ratio is as low as $0.2$, \method maintains a high accuracy in identifying novel categories, surpassing the best baseline by 2.45\%. This robust performance across diverse proportion settings highlights \method's stability and efficiency.





\section{Conclusion}
\label{sec:conclusion}
In this work, we introduce Event-Centric Generalized Category Discovery (EC-GCD), a more realistic and challenging variant of GCD characterized by complex event narratives, subjective class boundaries, and category imbalance. To address these challenges, we propose \method. We leverage LLMs to extract patterns from complex event narratives and refine them with labeled data to align with human criteria. The event patterns are utilized to adjust cluster assignments and cluster prototypes. To prevent dominant-class bias, we employ a ranking-filtering-mining strategy that enhances pattern diversity. We also contribute a new EC-GCD benchmark, Scam Report. Extensive experiments show that \method outperforms baselines on EC-GCD and generalizes well to base GCD benchmarks.

\bibliography{ref}

\begin{thebibliography}{10}

\bibitem{GCD}
Sagar Vaze, Kai Han, Andrea Vedaldi, and Andrew Zisserman.
\newblock Generalized category discovery.
\newblock In {\em {IEEE/CVF} Conference on Computer Vision and Pattern Recognition, {CVPR} 2022, New Orleans, LA, USA, June 18-24, 2022}, pages 7482--7491, 2022.

\bibitem{concept_gcd}
Nan Pu, Zhun Zhong, and Nicu Sebe.
\newblock Dynamic conceptional contrastive learning for generalized category discovery.
\newblock In {\em {IEEE/CVF} Conference on Computer Vision and Pattern Recognition, {CVPR} 2023, Vancouver, BC, Canada, June 17-24, 2023}, pages 7579--7588. {IEEE}, 2023.

\bibitem{Happy}
Shijie Ma, Fei Zhu, Zhun Zhong, Wenzhuo Liu, Xu{-}Yao Zhang, and Chenglin Liu.
\newblock Happy: {A} debiased learning framework for continual generalized category discovery.
\newblock In Amir Globersons, Lester Mackey, Danielle Belgrave, Angela Fan, Ulrich Paquet, Jakub~M. Tomczak, and Cheng Zhang, editors, {\em Advances in Neural Information Processing Systems 38: Annual Conference on Neural Information Processing Systems 2024, NeurIPS 2024, Vancouver, BC, Canada, December 10 - 15, 2024}, 2024.

\bibitem{MTP}
Yuwei Zhang, Haode Zhang, Li{-}Ming Zhan, Xiao{-}Ming Wu, and Albert Y.~S. Lam.
\newblock New intent discovery with pre-training and contrastive learning.
\newblock In Smaranda Muresan, Preslav Nakov, and Aline Villavicencio, editors, {\em Proceedings of the 60th Annual Meeting of the Association for Computational Linguistics (Volume 1: Long Papers), {ACL} 2022, Dublin, Ireland, May 22-27, 2022}, pages 256--269, 2022.

\bibitem{DPN}
Wenbin An, Feng Tian, Qinghua Zheng, Wei Ding, Qianying Wang, and Ping Chen.
\newblock Generalized category discovery with decoupled prototypical network.
\newblock In Brian Williams, Yiling Chen, and Jennifer Neville, editors, {\em Thirty-Seventh {AAAI} Conference on Artificial Intelligence, {AAAI} 2023, Thirty-Fifth Conference on Innovative Applications of Artificial Intelligence, {IAAI} 2023, Thirteenth Symposium on Educational Advances in Artificial Intelligence, {EAAI} 2023, Washington, DC, USA, February 7-14, 2023}, pages 12527--12535, 2023.

\bibitem{LOOP}
Wenbin An, Wenkai Shi, Feng Tian, Haonan Lin, Qianying Wang, Yaqiang Wu, Mingxiang Cai, Luyan Wang, Yan Chen, Haiping Zhu, and Ping Chen.
\newblock Generalized category discovery with large language models in the loop.
\newblock In Lun{-}Wei Ku, Andre Martins, and Vivek Srikumar, editors, {\em Findings of the Association for Computational Linguistics, {ACL} 2024, Bangkok, Thailand and virtual meeting, August 11-16, 2024}, pages 8653--8665, 2024.

\bibitem{TAN}
Wenbin An, Feng Tian, Wenkai Shi, Yan Chen, Yaqiang Wu, Qianying Wang, and Ping Chen.
\newblock Transfer and alignment network for generalized category discovery.
\newblock In Michael~J. Wooldridge, Jennifer~G. Dy, and Sriraam Natarajan, editors, {\em Thirty-Eighth {AAAI} Conference on Artificial Intelligence, {AAAI} 2024, Thirty-Sixth Conference on Innovative Applications of Artificial Intelligence, {IAAI} 2024, Fourteenth Symposium on Educational Advances in Artificial Intelligence, {EAAI} 2014, February 20-27, 2024, Vancouver, Canada}, pages 10856--10864, 2024.

\bibitem{ALUP}
Jinggui Liang, Lizi Liao, Hao Fei, Bobo Li, and Jing Jiang.
\newblock Actively learn from llms with uncertainty propagation for generalized category discovery.
\newblock In Kevin Duh, Helena G{\'{o}}mez{-}Adorno, and Steven Bethard, editors, {\em Proceedings of the 2024 Conference of the North American Chapter of the Association for Computational Linguistics: Human Language Technologies (Volume 1: Long Papers), {NAACL} 2024, Mexico City, Mexico, June 16-21, 2024}, pages 7845--7858, 2024.

\bibitem{GLEAN}
Henry~Peng Zou, Siffi Singh, Yi~Nian, Jianfeng He, Jason Cai, Saab Mansour, and Hang Su.
\newblock {GLEAN:} generalized category discovery with diverse and quality-enhanced {LLM} feedback.
\newblock {\em CoRR}, abs/2502.18414, 2025.

\bibitem{bank}
I{\~{n}}igo Casanueva, Tadas Temcinas, Daniela Gerz, Matthew Henderson, and Ivan Vulic.
\newblock Efficient intent detection with dual sentence encoders.
\newblock {\em CoRR}, abs/2003.04807, 2020.

\bibitem{stackoverflow}
Jiaming Xu, Peng Wang, Guanhua Tian, Bo~Xu, Jun Zhao, Fangyuan Wang, and Hongwei Hao.
\newblock Short text clustering via convolutional neural networks.
\newblock In Phil Blunsom, Shay~B. Cohen, Paramveer~S. Dhillon, and Percy Liang, editors, {\em Proceedings of the 1st Workshop on Vector Space Modeling for Natural Language Processing, VS@NAACL-HLT 2015, June 5, 2015, Denver, Colorado, {USA}}, pages 62--69, 2015.

\bibitem{clinc}
Stefan Larson, Anish Mahendran, Joseph~J. Peper, Christopher Clarke, Andrew Lee, Parker Hill, Jonathan~K. Kummerfeld, Kevin Leach, Michael~A. Laurenzano, Lingjia Tang, and Jason Mars.
\newblock An evaluation dataset for intent classification and out-of-scope prediction.
\newblock In Kentaro Inui, Jing Jiang, Vincent Ng, and Xiaojun Wan, editors, {\em Proceedings of the 2019 Conference on Empirical Methods in Natural Language Processing and the 9th International Joint Conference on Natural Language Processing, {EMNLP-IJCNLP} 2019, Hong Kong, China, November 3-7, 2019}, pages 1311--1316, 2019.

\bibitem{event-centric}
Muhao Chen, Hongming Zhang, Qiang Ning, Manling Li, Heng Ji, Kathleen McKeown, and Dan Roth.
\newblock Event-centric natural language processing.
\newblock In David Chiang and Min Zhang, editors, {\em Proceedings of the 59th Annual Meeting of the Association for Computational Linguistics and the 11th International Joint Conference on Natural Language Processing: Tutorial Abstracts}, pages 6--14, August 2021.

\bibitem{NCL}
Zhun Zhong, Enrico Fini, Subhankar Roy, Zhiming Luo, Elisa Ricci, and Nicu Sebe.
\newblock Neighborhood contrastive learning for novel class discovery.
\newblock In {\em Proceedings of the IEEE/CVF Conference on Computer Vision and Pattern Recognition (CVPR)}, pages 10867--10875, 2021.

\bibitem{DeCrisisMB}
Henry Zou, Yue Zhou, Weizhi Zhang, and Cornelia Caragea.
\newblock {D}e{C}risis{MB}: Debiased semi-supervised learning for crisis tweet classification via memory bank.
\newblock In Houda Bouamor, Juan Pino, and Kalika Bali, editors, {\em Findings of the Association for Computational Linguistics: EMNLP 2023}, pages 6104--6115, 2023.

\bibitem{mou}
Yutao Mou, Keqing He, Pei Wang, Yanan Wu, Jingang Wang, Wei Wu, and Weiran Xu.
\newblock Watch the neighbors: A unified k-nearest neighbor contrastive learning framework for {OOD} intent discovery.
\newblock In Yoav Goldberg, Zornitsa Kozareva, and Yue Zhang, editors, {\em Proceedings of the 2022 Conference on Empirical Methods in Natural Language Processing}, pages 1517--1529, 2022.

\bibitem{KTN}
Wenkai Shi, Wenbin An, Feng Tian, Yan Chen, Yaqiang Wu, Qianying Wang, and Ping Chen.
\newblock A unified knowledge transfer network for generalized category discovery.
\newblock In Michael~J. Wooldridge, Jennifer~G. Dy, and Sriraam Natarajan, editors, {\em Thirty-Eighth {AAAI} Conference on Artificial Intelligence, {AAAI} 2024, Thirty-Sixth Conference on Innovative Applications of Artificial Intelligence, {IAAI} 2024, Fourteenth Symposium on Educational Advances in Artificial Intelligence, {EAAI} 2014, February 20-27, 2024, Vancouver, Canada}, pages 18961--18969, 2024.

\bibitem{SimGCD}
Xin Wen, Bingchen Zhao, and Xiaojuan Qi.
\newblock Parametric classification for generalized category discovery: A baseline study.
\newblock In {\em 2023 IEEE/CVF International Conference on Computer Vision (ICCV)}, pages 16544--16554, 2023.

\bibitem{BaCon}
Jianhong Bai, Zuozhu Liu, Hualiang Wang, Ruizhe Chen, Lianrui Mu, Xiaomeng Li, Joey~Tianyi Zhou, YANG FENG, Jian Wu, and Haoji Hu.
\newblock Towards distribution-agnostic generalized category discovery.
\newblock In A.~Oh, T.~Naumann, A.~Globerson, K.~Saenko, M.~Hardt, and S.~Levine, editors, {\em Advances in Neural Information Processing Systems}, pages 58625--58647, 2023.

\bibitem{karamcheti}
Siddharth Karamcheti, Ranjay Krishna, Li~Fei-Fei, and Christopher Manning.
\newblock Mind your outliers! investigating the negative impact of outliers on active learning for visual question answering.
\newblock In Chengqing Zong, Fei Xia, Wenjie Li, and Roberto Navigli, editors, {\em Proceedings of the 59th Annual Meeting of the Association for Computational Linguistics and the 11th International Joint Conference on Natural Language Processing (Volume 1: Long Papers)}, pages 7265--7281, 2021.

\bibitem{MHPL}
Fan Wang, Zhongyi Han, Zhiyan Zhang, Rundong He, and Yilong Yin.
\newblock Mhpl: Minimum happy points learning for active source free domain adaptation.
\newblock In {\em 2023 IEEE/CVF Conference on Computer Vision and Pattern Recognition (CVPR)}, pages 20008--20018, 2023.

\bibitem{pmlr}
Yarin Gal, Riashat Islam, and Zoubin Ghahramani.
\newblock Deep {B}ayesian active learning with image data.
\newblock In Doina Precup and Yee~Whye Teh, editors, {\em Proceedings of the 34th International Conference on Machine Learning}, pages 1183--1192, 2017.

\bibitem{margatin}
Katerina Margatina, Loïc Barrault, and Nikolaos Aletras.
\newblock On the importance of effectively adapting pretrained language models for active learning.
\newblock {\em CoRR}, abs/2104.08320, 2022.

\bibitem{Citovsky}
Gui Citovsky, Giulia DeSalvo, Claudio Gentile, Lazaros Karydas, Anand Rajagopalan, Afshin Rostamizadeh, and Sanjiv Kumar.
\newblock Batch active learning at scale.
\newblock In M.~Ranzato, A.~Beygelzimer, Y.~Dauphin, P.S. Liang, and J.~Wortman Vaughan, editors, {\em Advances in Neural Information Processing Systems}, pages 11933--11944, 2021.

\bibitem{ACTUNE}
Dongyu Ru, Jiangtao Feng, Lin Qiu, Hao Zhou, Mingxuan Wang, Weinan Zhang, Yong Yu, and Lei Li.
\newblock Active sentence learning by adversarial uncertainty sampling in discrete space.
\newblock In Trevor Cohn, Yulan He, and Yang Liu, editors, {\em Findings of the Association for Computational Linguistics: EMNLP 2020}, pages 4908--4917, 2020.

\bibitem{DeepAL+}
Xueying Zhan, Qingzhong Wang, Kuan hao Huang, Haoyi Xiong, Dejing Dou, and Antoni~B. Chan.
\newblock A comparative survey of deep active learning.
\newblock {\em CoRR}, abs/2203.13450, 2022.

\bibitem{probabilistic}
Yunhua Zhou, Guofeng Quan, and Xipeng Qiu.
\newblock A probabilistic framework for discovering new intents.
\newblock In Anna Rogers, Jordan Boyd-Graber, and Naoaki Okazaki, editors, {\em Proceedings of the 61st Annual Meeting of the Association for Computational Linguistics (Volume 1: Long Papers)}, pages 3771--3784, 2023.

\bibitem{CsePL}
Jinggui Liang and Lizi Liao.
\newblock {C}luster{P}rompt: Cluster semantic enhanced prompt learning for new intent discovery.
\newblock In Houda Bouamor, Juan Pino, and Kalika Bali, editors, {\em Findings of the Association for Computational Linguistics: EMNLP 2023}, pages 10468--10481, 2023.

\bibitem{Hungarian}
Harold~W. Kuhn.
\newblock The hungarian method for the assignment problem.
\newblock In Michael J{\"{u}}nger, Thomas~M. Liebling, Denis Naddef, George~L. Nemhauser, William~R. Pulleyblank, Gerhard Reinelt, Giovanni Rinaldi, and Laurence~A. Wolsey, editors, {\em 50 Years of Integer Programming 1958-2008 - From the Early Years to the State-of-the-Art}, pages 29--47. 2010.

\bibitem{student-t}
Junyuan Xie, Ross~B. Girshick, and Ali Farhadi.
\newblock Unsupervised deep embedding for clustering analysis.
\newblock In Maria{-}Florina Balcan and Kilian~Q. Weinberger, editors, {\em Proceedings of the 33nd International Conference on Machine Learning, {ICML} 2016, New York City, NY, USA, June 19-24, 2016}, volume~48 of {\em {JMLR} Workshop and Conference Proceedings}, pages 478--487, 2016.

\bibitem{InfoNCE}
A{\"{a}}ron van~den Oord, Yazhe Li, and Oriol Vinyals.
\newblock Representation learning with contrastive predictive coding.
\newblock {\em CoRR}, abs/1807.03748, 2018.

\bibitem{prototype-learning}
Jake Snell, Kevin Swersky, and Richard~S. Zemel.
\newblock Prototypical networks for few-shot learning.
\newblock In Isabelle Guyon, Ulrike von Luxburg, Samy Bengio, Hanna~M. Wallach, Rob Fergus, S.~V.~N. Vishwanathan, and Roman Garnett, editors, {\em Advances in Neural Information Processing Systems 30: Annual Conference on Neural Information Processing Systems 2017, December 4-9, 2017, Long Beach, CA, {USA}}, pages 4077--4087, 2017.

\bibitem{ccl}
Chengjie Sun, Jie Ji, Boyue Shang, and Binguan Liu.
\newblock Overview of {CCL}23-eval task 6: Telecom network fraud case classification.
\newblock In {\em Proceedings of the 22nd Chinese National Conference on Computational Linguistics (Volume 3: Evaluations)}, pages 193--200, Harbin, China, August 2023. Chinese Information Processing Society of China.

\bibitem{h-score}
Kuniaki Saito and Kate Saenko.
\newblock Ovanet: One-vs-all network for universal domain adaptation.
\newblock In {\em 2021 {IEEE/CVF} International Conference on Computer Vision, {ICCV} 2021, Montreal, QC, Canada, October 10-17, 2021}, pages 8980--8989, 2021.

\bibitem{BERT}
Jacob Devlin, Ming{-}Wei Chang, Kenton Lee, and Kristina Toutanova.
\newblock {BERT:} pre-training of deep bidirectional transformers for language understanding.
\newblock In Jill Burstein, Christy Doran, and Thamar Solorio, editors, {\em Proceedings of the 2019 Conference of the North American Chapter of the Association for Computational Linguistics: Human Language Technologies, {NAACL-HLT} 2019, Minneapolis, MN, USA, June 2-7, 2019, Volume 1 (Long and Short Papers)}, pages 4171--4186, 2019.

\bibitem{qwen2.5}
An~Yang, Baosong Yang, Beichen Zhang, Binyuan Hui, Bo~Zheng, Bowen Yu, Chengyuan Li, Dayiheng Liu, Fei Huang, Haoran Wei, Huan Lin, Jian Yang, Jianhong Tu, Jianwei Zhang, Jianxin Yang, Jiaxi Yang, Jingren Zhou, Junyang Lin, Kai Dang, Keming Lu, Keqin Bao, Kexin Yang, Le~Yu, Mei Li, Mingfeng Xue, Pei Zhang, Qin Zhu, Rui Men, Runji Lin, Tianhao Li, Tingyu Xia, Xingzhang Ren, Xuancheng Ren, Yang Fan, Yang Su, Yichang Zhang, Yu~Wan, Yuqiong Liu, Zeyu Cui, Zhenru Zhang, and Zihan Qiu.
\newblock Qwen2.5 technical report.
\newblock {\em CoRR}, abs/2412.15115, 2024.

\end{thebibliography}
\bibliographystyle{unsrt}

\newpage
\appendix
\section{Limitations}
\label{sec:limitation}
While our method demonstrates strong performance on both EC-GCD and standard GCD benchmarks, it currently operates exclusively in the textual domain. The reliance on language models constrains its direct applicability to vision or multimodal settings. Extending our approach to these domains would require the integration of multimodal foundation models, which we leave for future work.

In addition, invoking large language models introduces non-negligible computational overhead, which may limit scalability in low-resource scenarios.

Lastly, while our pattern refinement step effectively captures annotator-defined decision boundaries for known classes, learning precise criteria for novel categories remains an open challenge. Future work could explore more adaptive mechanisms for inferring human-aligned criteria in truly unknown classes.

\section{Broader Impacts}
\label{sec:broader_impact}
Our construction of a scam report dataset as an EC-GCD benchmark carries significant potential benefits to both the research community and society. The primary positive impact lies in facilitating future research on EC-GCD, particularly on emerging scam classification and detection, which could enable enterprises and government agencies to more rapidly identify novel fraudulent schemes. This may lead to improved consumer protection mechanisms and more effective policy interventions against financial crimes. 

However, we recognize the potential risks of misuse, as malicious actors could theoretically exploit detailed scam descriptions to refine fraudulent strategies. To mitigate this concern, our dataset is currently undergoing an open-source review process and will only be released after rigorous desensitization. The final accessibility and permissible use cases will be determined based on the review outcomes, ensuring compliance with ethical guidelines and minimizing potential harm. 

\section{More Implementation Details}
\label{sec:implement}
\paragraph{Dataset Setup.} For the base GCD benchmarks (BANKING, StackOverflow, and CLINC), we follow An et al.~\cite{LOOP} by randomly selecting 25\% of the categories as novel and using 10\% of the data from each known category as labeled. For the EC-GCD datasets (Scam Report and Telecom Fraud Case), we select 35\% of the categories as novel and sample 10\% of the labeled data per known category. Dataset statistics are summarized in Table~\ref{tab:dataset_stats}. Additional experiments under varying known category ratios are provided in Figure~\ref{fig:kcr}.

\begin{table}[h]
\centering
\caption{Statistics of datasets. \(|\mathcal{Y}_k|\), \(|\mathcal{Y}_n|\), \(|\mathcal{D}^t|\), \(|\mathcal{D}^u|\) and \(|\mathcal{D}^t|\) represent the number of known categories, novel categories, labeled data, unlabeled data and testing data, respectively.}
\begin{tabular}{lcccccc}
\toprule
Dataset & \(|\mathcal{Y}_k|\) & \(|\mathcal{Y}_n|\) & \(|\mathcal{D}^l|\) & \(|\mathcal{D}^u|\) & \(|\mathcal{D}^t|\)  & License\\ \midrule
Scam Report & 14& 8& 724& 7,878& 1,076 & -\\
Telecom Fraud Case & 8 & 4 & 615& 7,272& 986 & unknown\\ \midrule
BANKING & 58 & 19 & 673 & 8,330 & 3,080 & CC-BY-4.0\\ 
StackOverflow & 15 & 5 & 1,350 & 16,650 & 1,000 & unknown\\ 
CLINC & 113 & 37 & 1,344 & 16,656 & 2,250 & unknown\\ \bottomrule
\end{tabular}
\label{tab:dataset_stats}
\end{table}

In addition, during the conduct of Scam Report, the average annotation consistency rate reached 91.8\%, and the category distribution is shown in Figure~\ref{fig:distribution}:
\begin{figure}[h]
    \centering
    \includegraphics[width=0.7\linewidth]{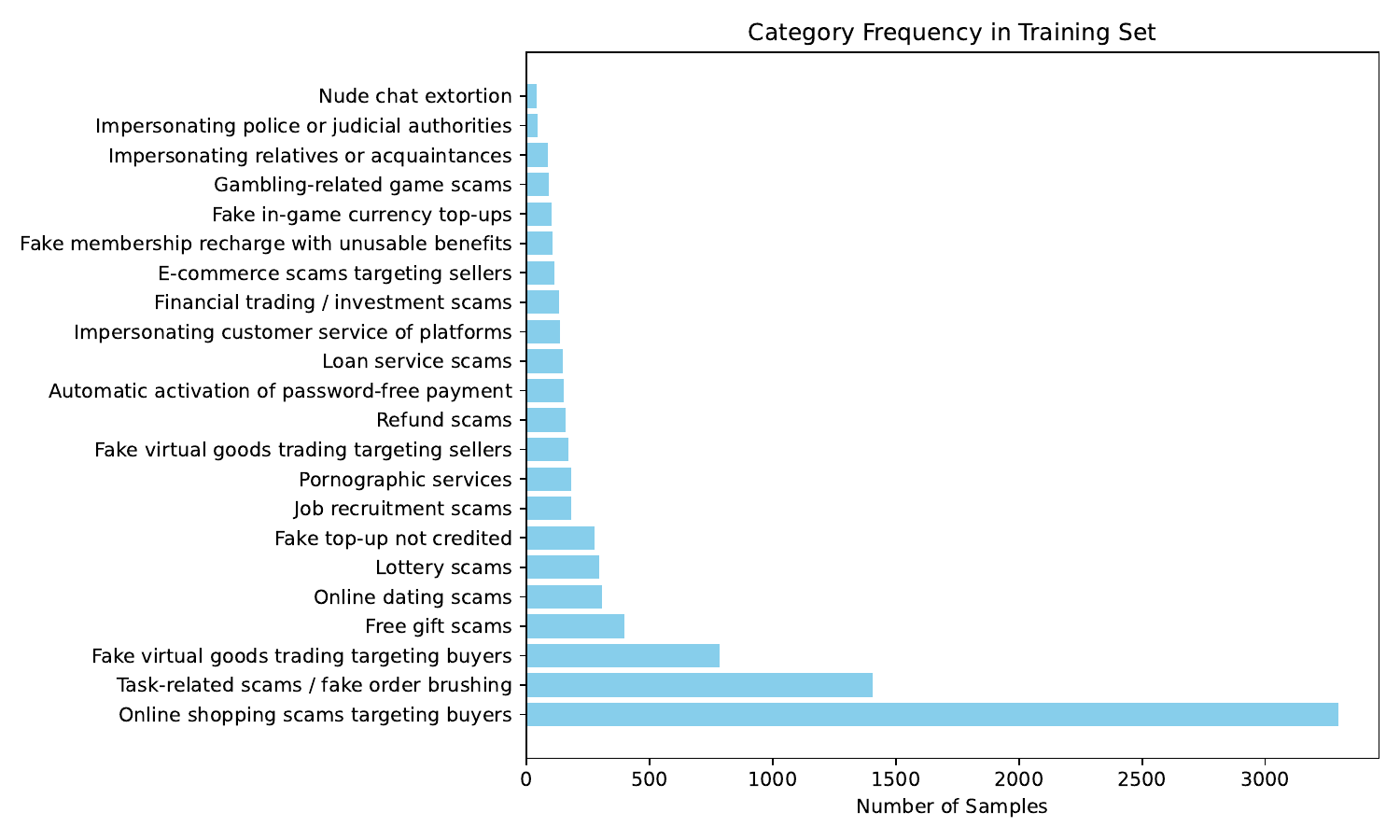}
    \caption{The category distribution of Scam Report's training set.}
    \label{fig:distribution}
\end{figure}

\paragraph{Evaluation Metrics}
\label{sec:metrics}
We use three standard evaluation metrics: Accuracy on known categories ($\text{ACC}_\text{K}$), Accuracy on novel categories ($\text{ACC}_\text{N}$), and H-score~\cite{h-score}. H-score is defined as the harmonic mean of $\text{ACC}_\text{K}$ and $\text{ACC}_\text{N}$:
\begin{equation}
    \text{H-score} = \frac{2ACC_K\cdot ACC_N}{ACC_K+ACC_N}.
\end{equation}

\paragraph{Hyperparameters}
A full list of default hyperparameters of our main experiment is provided in Table~\ref{tab:hyperparameter}. Our experiments are conducted on a single A100 GPU.

\begin{table}[h!]
\centering
\caption{Hyperparameter settings used in our experiments.}
\resizebox{\textwidth}{!}{%
\begin{tabular}{ll|ll}
\toprule
\textbf{Parameter} & \textbf{Value} & \textbf{Parameter} & \textbf{Value} \\
\midrule
Batch size & 32 & Temperature ($\tau$) & 0.07 \\
Learning rate & $1 \times 10^{-5}$ & Class centers' ratio ($\beta$) & 0.8 \\
Training epochs & 50 & Prototype retention weight ($\omega$) & 0.9 \\
High-confidence sample count ($k_{\text{high}}$) & 50 per class & K-means and LLM update interval & 5 \\
High-entropy sample count ($k_{\text{low}}$) & 500 in total & K-means runs per update & 5 \\
Compactness weight ($\sigma$) & 0.5 & Negative pairs per sample & 10 \\
Degrees of freedom ($\alpha$) & 1 & Transfer weight ($\rho$) & 1 \\
\bottomrule
\end{tabular}%
}
\label{tab:hyperparameter}
\end{table}

\UseRawInputEncoding
\lstset{
  language=Python,         
  basicstyle=\ttfamily\small,  
  keywordstyle=\color{blue},   
  stringstyle=\color{darkgreen},     
  commentstyle=\color{darkgreen},   
  numbers=left,                
  numberstyle=\tiny,           
  stepnumber=1,                
  breaklines=true,             
  frame=single,                
}

\section{More Ablation Studies on Different Parameters}
\label{sec:more_ablation}
\begin{figure}[h]
  \centering
  \includegraphics[width=\textwidth]{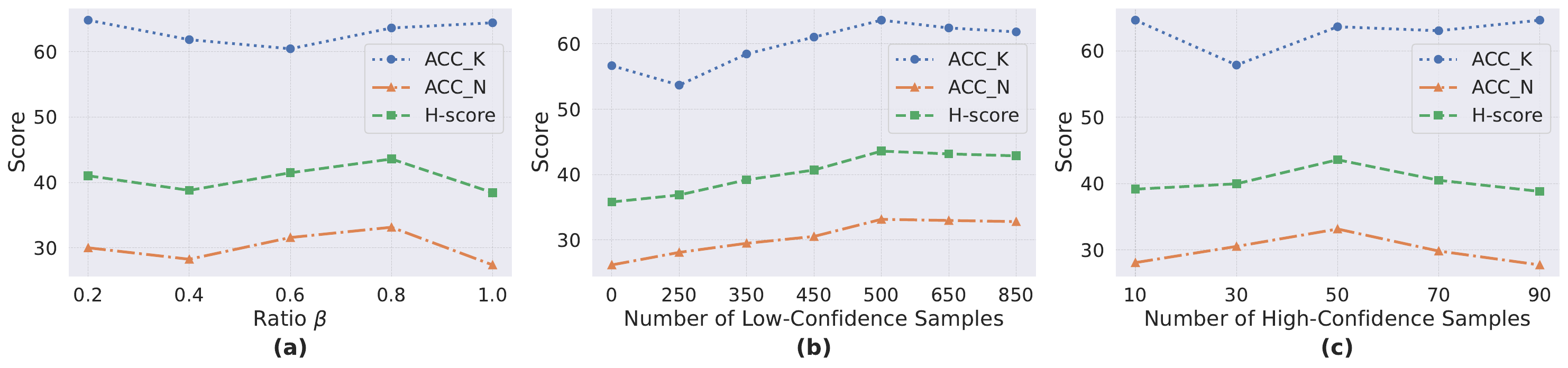}
  \caption{Impact of different parameters on \method's performance. (a) Ratio of the class center in prototype computation ($\beta$). (b) Number of high-confidence samples used for pattern generation. (c) Number of low-confidence samples selected for pseudo-label reassignment.}
  \label{fig:combine_sample_num_transfer_weight}
\end{figure}

\paragraph{The Ratio of Class Center in Prototype Calculation.}
As shown in Figure~\ref{fig:combine_sample_num_transfer_weight}(a), varying the ratio $\beta$ used for incorporating class centers in prototype computation significantly affects performance on both known and novel classes. A low ratio ($\beta=0.2$) emphasizes raw pattern embeddings and yields relatively high known-class accuracy ($\text{ACC}_\text{K}=64.81\%$), but performs poorly on novel classes ($\text{ACC}_\text{N}=30.02\%$), resulting in a low H-score (41.03\%). As $\beta$ increases, the known-class accuracy slightly drops, but novel-class performance gradually improves, peaking at $\beta=0.8$ with the highest H-score (43.59\%). This indicates that moderate inclusion of class centers (e.g., $\beta=0.8$) provides a better trade-off between preserving known-class structure and generalizing to novel categories. In contrast, full reliance on class centers ($\beta=1.0$) improves $\text{ACC}_\text{K}$ (64.41\%) but causes a significant drop in novel-class accuracy (27.40\%), leading to the lowest H-score (38.45\%). Overall, $\beta=0.8$ achieves the best balance between generalization and stability.

\paragraph{The Number of Low-Confidence Samples.}

As shown in Figure \ref{fig:combine_sample_num_transfer_weight}(b), to investigate the sensitivity of \method to the number of high-entropy samples, we perform an ablation by varying the number of top-entropy samples selected per epoch before intersecting with the unstable sample set. Specifically, we experiment with selecting $\{0, 250, 350, 450, 500, 650, 850\}$ high-entropy samples. The results reveal that selecting too few low-confidence samples (e.g., 0 or 250) limits the intersection size, potentially overlooking mislabeled samples that could benefit from reassignment. Conversely, selecting too many (e.g., 850) may increase noise by including samples with ambiguous semantics, thereby weakening the precision of reassignment. The best performance is typically observed at intermediate settings (e.g., 500), which strike a balance between sample coverage and reliability.

\paragraph{The Number of High-Confidence Samples}

As shown in Figure \ref{fig:combine_sample_num_transfer_weight}(c), the identification of high-confidence samples is a key step in our framework for guiding the generation of reliable patterns. To assess the impact of high-confidence sample quantity on model performance, we conduct an ablation study by varying the number of top low-entropy and high-centrality samples selected per cluster before intersection. Specifically, we test values in the set $\{10, 30, 50, 70, 90\}$. The results indicate that model performance initially improves as sample number increases, peaking at 50. This indicates that a moderate number of high-confidence samples provide the LLM with enough semantic evidence to construct expressive and category-specific patterns. When sample number is too small (e.g., 10), the intersection set becomes sparse and insufficient for meaningful pattern induction. On the other hand, larger values (e.g., 70 or 90) expand the candidate pool but introduce samples with lower confidence, leading to less precise patterns and decreased overall performance. These results confirm the need to balance representational adequacy and noise suppression when selecting high-confidence samples.

\section{The Example of Sample Matching}
\label{sec:example_matching}
The following shows a sample where the initial pseudo-label was wrong and then corrected during the sample matching phase. All contexts have been translated into English for demonstration purposes, and sensitive information such as platform names or public account names has been replaced with 'XX'.

\textbf{Sample information:}
\begin{table}[h]
    \centering
    \begin{tabular}{rp{10cm}} \toprule
       \textbf{Sample:}  &  "In CrossFire, the Phantom God sound card - he contacted me on the XX platform and said I could buy the Phantom God sound card for 488 yuan. I then added his contact information. He sent me a payment code, and I transferred the money. In total, I made 15 transfers amounting to 2,880 yuan. However, he never shipped the item and has not refunded the money."\\
       \textbf{Initialized pseudo-label:}  & Online shopping scam \\
       \textbf{Ground-truth label:} & Deceptive transaction related to virtual game items \\\bottomrule
    \end{tabular}
    \label{tab:my_label}
\end{table}

Sample Matching Process (The unimportant content is omitted.):
\begin{lstlisting}
Input = '''
Task Objective:
1. Analyze the user-provided text samples.
2. Match each sample to a predefined category from the provided category set.
3. Strictly follow the required output format.

Category Set:
1: Fraudulent Product Transactions Scammers post attractive product listings on social or shopping platforms (such as Xianyu or Xiaohongshu). When a victim shows interest in purchasing, the scammer guides them to complete the transaction offline through various means (e.g., WeChat, QQ), and requests payment in advance. Once the money is received, the scammer provides a fake tracking number or logistics information, then ceases all communication and never delivers the product or service.
2: Fake Cashback from Order-Brushing Scams: ... 
3: Game Account/Boosting Service Scams Scammers contact victims through social or gaming platforms, offering game accounts, in-game items, or boosting services at prices lower than the market rate. They gradually persuade the victim to pay a deposit or the full amount. After receiving the payment, scammers either fabricate excuses (e.g., account issues, system errors) to demand additional payments or directly delete the victim from their contacts and go offline, resulting in financial loss. To make the scam harder to trace, scammers often ask for payment through multiple methods such as QQ Red Packets, Alipay, or WeChat Red Packets.
4: Discounted Mobile Credit Top-Up Scams: ... 
...

Text Samples to be Classified:
"In CrossFire, the Phantom God sound card—he contacted me on the XX platform and said I could buy the Phantom God sound card for 488 yuan. I then added his contact information. He sent me a payment code, and I transferred the money. In total, I made 15 transfers amounting to 2,880 yuan. However, he never shipped the item and has not refunded the money."

Let me know if you'd like it rewritten more formally or tailored for a specific use, like a police report or a customer support complaint.


Classification Rules:
1. Compare each text sample against the categories. A sample can only be assigned to a category if it fully matches the category definition and satisfies all of its core required features.
2. If a sample matches multiple categories, select the one with the highest degree of matching.
3. If any core feature is missing, classify the sample as a "New Category".
 Output Format Requirements:

Please strictly return the result in JSON format (do not use markdown formatting), including the following fields:
```json
{{
    "results": [
        {{
            "Index": Original text sample index,
            "Assigned Category Index": Category index or "New Category",
            "Matching Justification": An explanation of which core required features and auxiliary features were matched. The sample must satisfy all core required features to be assigned.
        }},
        ...
    ]
}}
```
'''


Output =
'''
[
  {
    "Index": 1,
    "SchemeCategoryIndex": 3,
    "MatchingBasis": "The report describes how the victim was approached via a social platform (XX) and a game trading platform, lured by a game item (Phantom God sound card) offered below market price. The scammer gradually induced the victim to make multiple payments (15 times totaling 2,880 yuan). After the payments were completed, the scammer did not deliver the item nor issue a refund. This aligns with the core criteria of a game account/boosting service scam: the scammer contacts the victim through social or game trading platforms, uses underpriced game items as bait, gradually guides the victim to make payments, and ultimately fails to provide the product or service."
  }
]
'''

\end{lstlisting}

\section{The Example of Pattern Generation}
\label{sec:example_generation}
The following is an example of pattern generation. All contexts have been translated into English for demonstration purposes, and sensitive information such as platform names or public account names has been replaced with 'XX'.
\begin{lstlisting}
Input = '''
Task Objective: Analyze the scam report to identify major scam patterns.

Output Requirements (do not use markdown format):

<FORMAT>
1. Scam Method Analysis:  
   Break down each report into key elements:  
   - Identity of the scammer (e.g., impersonating government, company, acquaintance, etc.)  
   - Keywords in the scam dialogue  
   - Characteristics of the fund flow  
   - Technical methods used  

2. Scam Type Statistics:
   [Show the number of reports per scam type, their proportions, and typical features; identify the scam type with the highest number of reports]

3. Summary of Main Scam Type Patterns:
   [Scam Type Name]: [Description of the typical scam flow/process for this type]

4. List of report numbers belonging to the main scam types identified in steps 2 and 3:
   [
   {
   "Report Number": number,
   "Basis": "..."
   },
   {
   "Report Number": number,
   "Basis": "..."
   },
   ... // Each element corresponds to a report number and its rationale
   ]

   </FORMAT>

Report Information:
1: I saw loan information on a website via my phone browser. After learning the details, I borrowed 10,000 RMB with a monthly interest of 60 RMB, which seemed reasonable. The loan requirements were an ID card front and back, and depositing 3,000 RMB into a bank card. I repeatedly confirmed that the 3,000 RMB would not be withdrawn before depositing. After I deposited, when I was about to proceed, they automatically took the money. I felt scammed, but they denied it, saying my transaction history was insufficient and I needed to transfer money back and forth via XX frequently for loan approval. After transferring all, they said my credit score was low and I had to deposit another 5,000 RMB. I refused, and they blocked me. I didn’t get the loan and lost 3,000 RMB. Although I feel the money is unrecoverable, I want to report it to prevent others from being scammed. The conversations were also on a corporate XX, but after being blocked, all chat records disappeared.

2: Initially, I wanted to borrow 70,000 RMB. After I made the request, they asked me to transfer 889 RMB to XX as a quota fee and said the loan would be disbursed within 20 minutes after paying. Later, they said the loan couldn’t be released, and I needed to top up 3,889 RMB transaction volume. I transferred 2,043 RMB to XX. They said the loan would be disbursed in 15 minutes. After 15 minutes, when I inquired, they said my account had problems and couldn’t release the loan. When I asked for a refund, they said the process had started and it was non-refundable. Then I realized I was scammed.

3: On the XX platform, they contacted me promising 100% loan approval. Then I added an XX friend and was told I only needed to pay 205.76 RMB as an account opening fee. The first application failed, so I paid another 205.76 RMB for reopening, totaling 411.52 RMB. Then they said that because I reapplied, I only needed to pay 300 RMB to get the loan directly. That’s when I realized it was a scam. When I asked for a refund, they ignored me.

4: This scam starts by using XX and other third-party apps to lure users to loan services. On XX, there’s a public account called XX, which offers loans via third-party apps. When the process moves from review to loan disbursement, users are asked to pay fees. After payment, they recommend another app to proceed again, requiring more fees. When users try to withdraw, more fees are requested. The cycle repeats with escalating fees, misleading users to invest but never releasing the loan.

5: Due to a personal urgent need, I searched for loan apps online. After registration, it showed that my application for 50,000 RMB was successful. The customer service said I needed to verify funds, requiring 30%-50% of the principal on my bank card. I first withdrew 15,000 RMB to reach 30%, then I was told I needed 50%, so I withdrew another 10,000 RMB. These funds were all taken by the scammer through forms they sent me, with transfers of 500 RMB each in Alipay group transfers, totaling more than 30 transactions.

6: The scammer claimed to be from XX, offering personal loans, required users to deposit 9,000 RMB into their bank card, and asked for mobile SMS verification codes to withdraw 5,000 RMB from the card. Then, they asked users to transfer 4,000 RMB via XX. Ultimately, the loan was not granted.

7: On the XX platform, I was told loans could be approved. After adding them, they said they don’t scam but require upfront fees. I paid about 100 RMB initially. Then they kept asking for more transfers, first 320 RMB as a promissory note fee, then 100 RMB because I didn’t add remarks, and then 220 RMB because the previous transfer failed.
'''

Output = '''
1. Scam Method Analysis:
   * Report 1: Scammer identity (posing as a loan institution); key phrases (loan, fund verification, transaction volume, comprehensive score); fund flow characteristics (via bank card, Alipay transfers); technical methods (through mobile browser website, corporate xx)
   * Report 2: Scammer identity (posing as a loan institution); key phrases (loan, quota fee, supplementing transaction volume); fund flow characteristics (transfers via xx); technical methods (via xx)
   * Report 3: Scammer identity (posing as a loan institution); key phrases (loan, account opening fee, re-opening account); fund flow characteristics (xx transfers); technical methods (via xx, xx)
   * Report 4: Scammer identity (posing as a loan institution); key phrases (loan, payment fees, escalating fees); fund flow characteristics (via xx public account, third-party apps); technical methods (via xx, xx public account, third-party apps)
   * Report 5: Scammer identity (posing as a loan institution); key phrases (loan, fund verification, group forwarding); fund flow characteristics (via xx group forwarding); technical methods (via online search for loan software, xx)
   * Report 6: Scammer identity (posing as xx); key phrases (personal loan, fund verification, SMS verification code); fund flow characteristics (via bank card, xx transfers); technical methods (via internet)
   * Report 7: Scammer identity (posing as a loan institution); key phrases (loan, promissory note fee, transfer with remarks); fund flow characteristics (via transfers); technical methods (via xx, transfers)

2. Scam Type Statistics:
   The scam type with the highest number of reports is "Loan Scam," with all 7 reports, accounting for 100%. Typical features include impersonating official loan institutions, requiring victims to pay various fees (such as fund verification fees, account opening fees, quota fees), and exploiting online platforms (such as mobile browsers, social media, third-party apps) to conduct scams.

3. Summary of Main Scam Type Patterns:
   Loan Scam: Scammers first post fake loan offers on online platforms (such as xx, xx) to attract victims. After contact, they impersonate official loan institutions and demand various fees (e.g., fund verification, account opening, quota fees). After victims pay, scammers use excuses (such as insufficient transaction volume or low comprehensive score) to demand higher fees repeatedly, never releasing the loan, and ultimately block the victims.

4. List of report numbers belonging to the main scam types identified in steps 2 and 3:
   [
   {
   "Report Number": 1,
   "Basis": "Loan scam, requiring payment of fund verification fee"
   },
   {
   "Report Number": 2,
   "Basis": "Loan scam, requiring quota fee payment and supplementing transaction volume"
   },
   {
   "Report Number": 3,
   "Basis": "Loan scam, requiring payment of account opening fees"
   },
   {
   "Report Number": 4,
   "Basis": "Loan scam, requiring payment through third-party apps"
   },
   {
   "Report Number": 5,
   "Basis": "Loan scam, requiring fund verification fee, and scamming via group forwarding"
   },
   {
   "Report Number": 6,
   "Basis": "Loan scam, impersonating Ping An Bank and requiring a fund verification fee"
   },
   {
   "Report Number": 7,
   "Basis": "Loan scam, requiring payment of promissory note and other fees"
   }
   ]
'''
\end{lstlisting}

\end{document}